\documentclass[sigconf, nonacm]{acmart}

\AtBeginDocument{%
  }

\setcopyright{acmlicensed}
\copyrightyear{2026}
\acmYear{2026}
\acmDOI{XXXXXXX.XXXXXXX}
\acmConference[ACM MM]{Make sure to enter the correct
  conference title from your rights confirmation email}{November 10–14, 2026}{Rio de Janeiro, Brazil}
\acmISBN{978-1-4503-XXXX-X/2018/06}

\usepackage{multirow}

\definecolor{secondcolor}{RGB}{220,230,240}
\definecolor{firstcolor}{RGB}{241,220,219}
\usepackage{xspace}
\newcommand{\ours}{\textsl{ReDiffuse}\xspace}
\usepackage{enumerate}

\usepackage[table]{xcolor}
\newcommand{\graytext}[1]{{\scriptsize{\textcolor{gray}{#1}}}}

\begin{document}

\title{ReDiffuse: Rotation Equivariant Diffusion Model for Multi-focus Image Fusion}

\author{Bo Li}
\email{libo0305@stu.xjtu.edu.cn}

\affiliation{%
  \institution{School of Computer Science and Technology, Xi'an Jiaotong University}
  \city{Xi'an}
  \state{Shaanxi}
  \country{China}
}

\author{Tingting Bao}
\email{baotting@stu.xjtu.edu.cn}
\affiliation{%
  \institution{School of Computer Science and Technology, Xi'an Jiaotong University}
  \city{Xi'an}
  \country{Shaanxi}
  \country{China}
  }

\author{Lingling Zhang}
\authornote{Corresponding author}
\email{zhanglling@xjtu.edu.cn}
\affiliation{%
  \institution{School of Computer Science and Technology, Xi'an Jiaotong University}
  \city{Xi'an}
  \country{Shaanxi}
  \country{China}
  }

\author{Weiping Fu}
\email{fuweiping@stu.xjtu.edu.cn}
\affiliation{%
  \institution{School of Computer Science and Technology, Xi'an Jiaotong University}
  \city{Xi'an}
  \country{Shaanxi}
  \country{China}
  }

\author{Yaxian Wang}
\email{wyx1566@chd.edu.cn}
\affiliation{%
  \institution{School of Information Engineering, Chang'an University}
  \city{Xi'an}
  \country{Shaanxi}
  \country{China}
  }

\author{Jun Liu}
\email{liukeen@xjtu.edu.cn}
\affiliation{%
  \institution{School of Computer Science and Technology, Xi'an Jiaotong University}
  \city{Xi'an}
  \country{Shaanxi}
  \country{China}
  }

\renewcommand{\shortauthors}{}
\renewcommand{\shorttitle}{}
\begin{abstract}
Diffusion models have achieved impressive performance on multi-focus image fusion (MFIF).
However, a key challenge in applying diffusion models to the ill-posed MFIF problem is that defocus blur can make common symmetric geometric structures (e.g., textures and edges) appear warped and deformed, often leading to unexpected artifacts in the fused images.
Therefore, embedding rotation equivariance into diffusion networks is essential, as it enables the fusion results to faithfully preserve the original orientation and structural consistency of geometric patterns underlying the input images.
Motivated by this, we propose \ours, a rotation-equivariant diffusion model for MFIF.
Specifically, we carefully construct the basic diffusion architectures to achieve end-to-end rotation equivariance.
We also provide a rigorous theoretical analysis to evaluate its intrinsic equivariance error, demonstrating the validity of embedding equivariance structures.
\ours is comprehensively evaluated against various MFIF methods across four datasets (Lytro, MFFW, MFI-WHU, and Road-MF).
Results demonstrate that \ours achieves competitive performance, with improvements of 0.28–6.64\% across six evaluation metrics. 
The code is available at \url{https://github.com/MorvanLi/ReDiffuse}.
\end{abstract}



\maketitle

\section{Introduction}
Due to the limited depth of field (DoF) of imaging devices, it is difficult to capture all regions of a scene in sharp focus within a single exposure.
Multi-focus image fusion (MFIF)~\cite{zhang2021deep} addresses this problem by integrating multiple images captured at different focal planes, combining their focused regions into a single all-in-focus image.
Such fused images provide more complete and reliable visual information, which is beneficial for downstream tasks such as detection~\cite{wang2021discriminative} and segmentation~\cite{liu2021searching}.
As a result, MFIF has long been considered an important topic in computer vision~\cite{11208806}.

\begin{figure}[t]
  \begin{center}
    \centerline{\includegraphics[width=\linewidth, keepaspectratio]{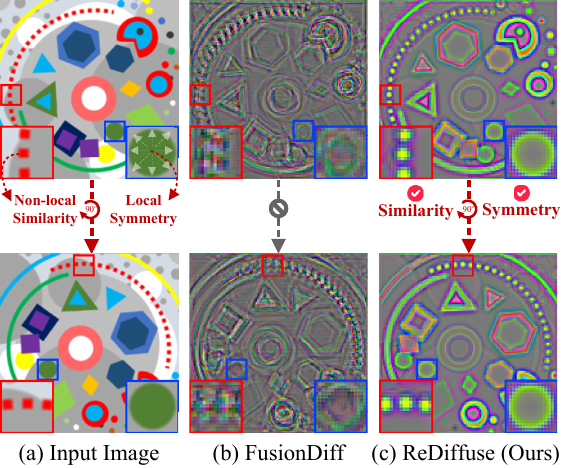}}
    \caption{
    (a) is one defocused source image in MFIF containing typical geometric structures.
    (b) and (c) show fused results of the FusionDiff~\cite{li2024fusiondiff} and our proposed \ours, respectively.
    Even under defocus, rotation equivariance enables \ours to better preserve non-local directional similarity and local isotropic symmetry, providing clear visual evidence of its benefits for the MFIF task.
    }
    \label{fig:Intro-1}
  \end{center}
  \vspace{-2em}
\end{figure}

Over the past decade, deep learning has substantially advanced MFIF~\cite{liu2020multi}, shifting fusion strategies from handcrafted designs to end-to-end learned frameworks.
Among generative methods, Generative Adversarial Network (GAN)-based approaches~\cite{huang2020generative,guo2019fusegan} have been explored for image fusion, but their training instability and mode collapse often limit reconstruction fidelity for complex image details.
Recently, diffusion models~\cite{ho2020denoising} have emerged as a powerful alternative generative paradigm.
Benefiting from a stable likelihood-based training objective, they are able to produce high-fidelity images and have shown growing potential in image fusion tasks~\cite{he2025dm}.

Despite achieving significant progress, directly applying diffusion models to the ill-posed MFIF problem remains challenging. 
In MFIF, source images are captured under different focal settings, where defocus blur often degrades edges, textures, and shape boundaries.
Such degradation weakens the geometric cues required for reliable cross-source correspondence and makes it difficult for diffusion models to progressively reconstruct geometrically consistent content during conditional denoising.
As a result, existing diffusion-based MFIF methods~\cite{li2024fusiondiff} can still produce noticeable artifacts in fused images, often manifested as warped edges, distorted textures, and deformed structures, as shown in \figurename~\ref{fig:Intro-1}(b).

Therefore, diffusion-based MFIF methods require more effective geometric priors to be embedded into their network architectures. 
It is worth noting that most existing methods~\cite{11297852,cao2024conditional,yi2024diff,tang2025mask} rely on heuristic image processing modules, overlooking the critical issue of maintaining structural consistency during the diffusion process.
Incorporating rotation equivariance~\cite{cohen2016group,worrall2017harmonic} into the diffusion architecture is particularly suitable for MFIF, as defocus blur weakens local geometric features, making structural recovery from source images captured at different focal settings increasingly dependent on consistent modeling of shared structures across orientations.
Rotation equivariance provides exactly such a geometric inductive bias by producing consistent responses to similarly structured patterns across different orientations, thereby helping preserve the orientation consistency and structural integrity of underlying geometric patterns during denoising.

To provide an intuitive illustration, we select a defocused source image containing typical geometric structures.
Rotation equivariance benefits the fusion performance of diffusion models in two key aspects.
First, embedding rotation equivariance helps diffusion models better preserve non-local directional similarity, as evidenced by the red-boxed comparison between \figurename~\ref{fig:Intro-1}(b) and \figurename~\ref{fig:Intro-1}(c).
This is because rotation equivariance leverages similarities across different directions in non-local regions of the image, ensuring that image features like small squares (with local similarity along smoothly varying curve directions) are more faithfully recovered in fused outputs.
Second, rotation equivariance also contributes to maintaining local isotropic symmetry, as highlighted by the blue box in \figurename~\ref{fig:Intro-1}(c). 
This property ensures that features with similar characteristics in all directions (e.g., isotropic circles) are reconstructed consistently, without introducing directional artifacts.

\begin{figure}[t]
  \begin{center}
    \centerline{\includegraphics[width=\linewidth, keepaspectratio]{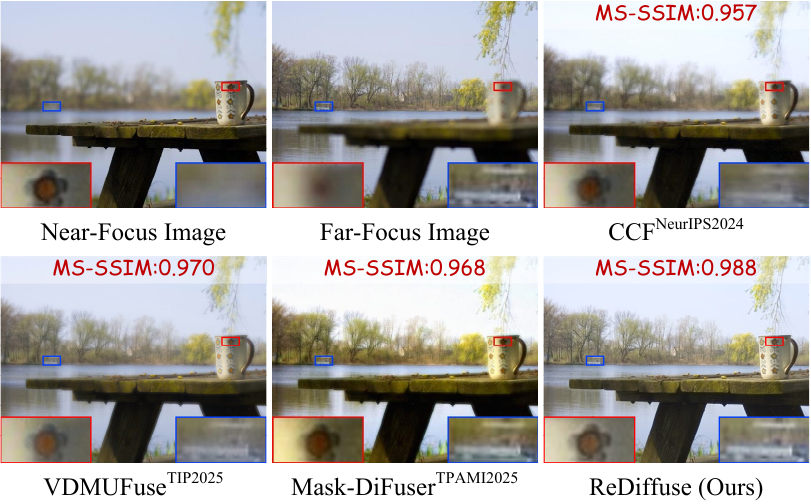}}
    \caption{
     Fusion results generated by three diffusion fusion methods and our \ours. 
     Our \ours achieves superior structural consistency and artifact suppression.
    }
    \label{fig:Intro-3}
  \end{center}
  \vspace{-1.5em}
\end{figure}

However, designing an MFIF diffusion architecture that intrinsically supports rotation equivariance remains challenging. 
Diffusion models typically adopt a U-Net backbone that integrates multiple modules with complex operators, such as upsampling and downsampling, which often disrupt rotation equivariance on discrete image grids.
This calls for a systematic redesign of the diffusion architecture to preserve rotation-equivariant representations throughout the denoising process. 
To this end, we propose a novel diffusion framework for MFIF, termed \ours.
The key lies in incorporating rotation equivariance into the diffusion network under a strict equivariance error guarantee.
Compared with existing methods, \ours effectively mitigates fusion artifacts caused by structural degradation, as shown in \figurename~\ref{fig:Intro-3}.

The key contributions include:
\vspace{-0.4em}
\begin{enumerate}
  \item[(1)] \textbf{Rotation Equivariance Technique.} We propose a rotation-equivariant diffusion MFIF method, termed \ours, which maintains approximate end-to-end rotational equivariance throughout the entire network.
  \item[(2)] \textbf{Theoretical Support.} We provide a comprehensive theoretical analysis of the intrinsic equivariance error of the entire proposed \ours network, including its upsampling, downsampling and normalization modules, demonstrating
its inherent nature of embedding such expected rotation equivariance structure.
  \item[(3)] \textbf{Performance Gains and Lightweight.} 
  We validate the superiority of the proposed \ours on four MFIF datasets. Extensive experiments show that our \ours improves performance by 0.28--6.64\% across six evaluation metrics. 
  Moreover, thanks to more efficient parameter sharing, it reduces the parameter count from 26.91M to 7.55M compared with the non-equivariant backbone.
\end{enumerate}

\section{Related Work}
\subsection{DL-based MFIF Methods}
\textbf{Decision-map method.}
These methods formulate MFIF as a classification problem, where deep neural networks are trained to predict decision maps that distinguish in-focus from out-of-focus regions and guide the fusion process~\cite{li2020drpl}.
To improve segmentation accuracy, various strategies have been introduced, including attention mechanisms~\cite{xiao2020global}, multi-scale architectures~\cite{liu2021multiscale}, Mamba-based modeling~~\cite{jin2025combining}, edge-aware designs~\cite{wang2023multi}, and the incorporation of physical priors~\cite{quan2025multi}.
However, under severe defocus conditions, these approaches often suffer from notable performance degradation, resulting in blurred boundaries and inaccurate region classification.

\textbf{End-to-end method.}
These methods employ end-to-end deep neural networks to directly map source images to fused outputs, typically adopting encoder–decoder architectures.
Representative approaches include CNNs~\cite{xu2020u2fusion,zhang2020rethinking,zhang2020ifcnn}, GANs~\cite{guo2019fusegan}, and Transformer-based architectures~\cite{ma2022swinfusion}.
Recently, diffusion models~\cite{li2024fusiondiff} have demonstrated outstanding performance in MFIF.
They often struggle to maintain the spatial structural consistency of source images, inevitably resulting in artifacts in the fused results.

\subsection{Rotation Equivariance}
Rotation equivariance ensures that input rotations induce consistent and predictable responses in the feature space.
Since local structures in multi-focus images (e.g., edges and textures) exhibit strong geometric consistency, incorporating rotation equivariance facilitates feature alignment during fusion.
Early filter parameterization methods~\cite{weiler2019general,weiler2018learning} suffer from limited expressive capacity and approximation accuracy, hindering their ability to model continuous rotations and thus limiting their performance on MFIF tasks.
Recent Fourier-series-based parameterizations (B-Conv)~\cite{xie2025rotation}, achieve strict rotational equivariance in the continuous domain.
However, theoretical guarantees for enforcing rotation equivariance within diffusion models remain largely unexplored.
Our work \ours provides a rigorous theoretical guarantee of rotation equivariance for diffusion models.

\subsection{Diffusion Model}
Diffusion models have attracted increasing attention in MFIF due to their powerful image generation capabilities and clear probabilistic modeling~\cite{Zhao_2023_ICCV,shi2024vdmufusion,tang2025mask}.
Despite achieving notable improvements, diffusion models still face challenges in the ill-posed MFIF setting, where defocus blur can distort symmetric geometric structures (e.g., textures and edges), resulting in artifacts in the fused images.
Unlike existing studies, our \ours embeds rotation equivariance into the diffusion framework, guiding the model to generate fusion results that more faithfully preserve the original directional consistency and structural integrity of underlying geometric patterns.

\section{Method}
\subsection{Problem Overview}
\begin{figure}[t]
  \begin{center}
    \centerline{\includegraphics[width=\linewidth, keepaspectratio]{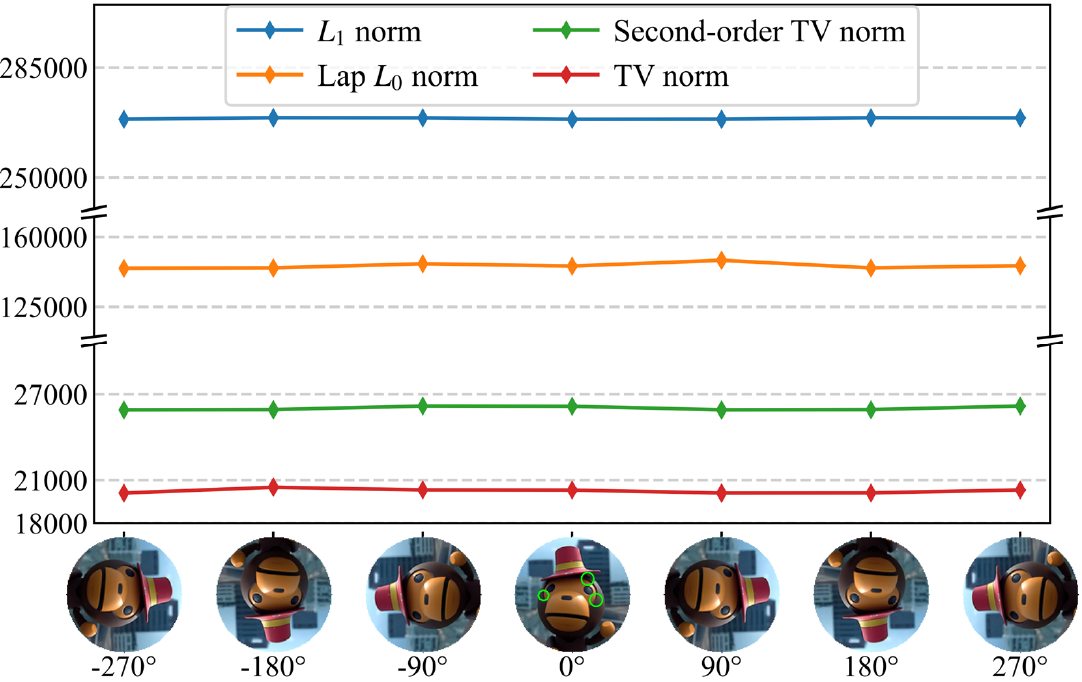}}
    \caption{
     Four typical conventional regularization values for near-focus images from the Lytro dataset at different rotation angles. 
     The circled regions show that many local structures at different orientations share similar geometric properties.
    }
    \label{fig:rotation-image}
  \end{center}
  \vspace{-1.5em}
\end{figure}

MFIF aims to integrate complementary in-focus regions from multiple source images into a single all-in-focus result.
Although diffusion models exhibit strong reconstruction ability, defocus blur often degrades local geometric structures, leading to ambiguous or inconsistent orientation cues during denoising.
Such inconsistencies may gradually accumulate over diffusion steps, eventually leading to structural artifacts in the fused image and thereby imposing higher demands on direction-consistent structural representations.
In MFIF, many local structures, although appearing at different orientations, often share similar geometric properties.
Accordingly, common image priors such as non-local self-similarity and low-rankness typically remain stable under rotation transformations, as illustrated in \figurename~\ref{fig:rotation-image}.
This suggests that introducing rotation-equivariant geometric priors into the diffusion backbone can help enhance geometric consistency and stabilize structural reconstruction.
Based on this idea, we propose a rotation-equivariant diffusion framework for MFIF, termed \ours.
The overall pipeline is illustrated in \figurename~\ref{fig:framewrk}.

\subsection{Rotation-Equivariant Diffusion Framework}
We first describe the diffusion-based formulation adopted in \ours, including the forward and reverse processes and the associated training objective. 
We then introduce the rotation-equivariant U-Net and provide a theoretical analysis of its equivariance properties.

\textbf{Forward Diffusion Process.}
The forward diffusion is a Markov chain that gradually perturbs an initial sample $F_0$ with Gaussian noise, yielding $F_T \sim \mathcal{N}(0,\mathbf{I})$.
The total number of diffusion steps $T$ is set to 2000.
The forward transition is defined as
\begin{equation}
q(F_t \mid F_{t-1})
=
\mathcal{N}\!\left(F_t;\sqrt{1-\beta_t}\,F_{t-1},\, \beta_t \mathbf{I} \right),
\end{equation}
where $\beta_t \in (0,1)$ is the noise schedule.

Note that $F_t$ can be directly obtained from $F_0$ by 
\begin{equation}
q(F_t \mid F_0)
=
\mathcal{N}\!\left(F_t;\sqrt{\bar{\alpha}_t}\,F_0,\,(1-\bar{\alpha}_t)\mathbf{I}\right),
\end{equation}
where $\alpha_t=1-\beta_t$ and $\bar{\alpha}_t=\prod_{s=1}^t\alpha_s$.

Using the reparameterization trick, we can obtain
\begin{equation}
F_t
=
\sqrt{\bar{\alpha}_t}\,F_0
+
\sqrt{1-\bar{\alpha}_t}\,\epsilon,
\quad \epsilon \sim \mathcal{N}(0,\mathbf{I}).
\label{eq:reparam}
\end{equation}

\begin{figure*}
\centering
  \includegraphics[width=\linewidth]{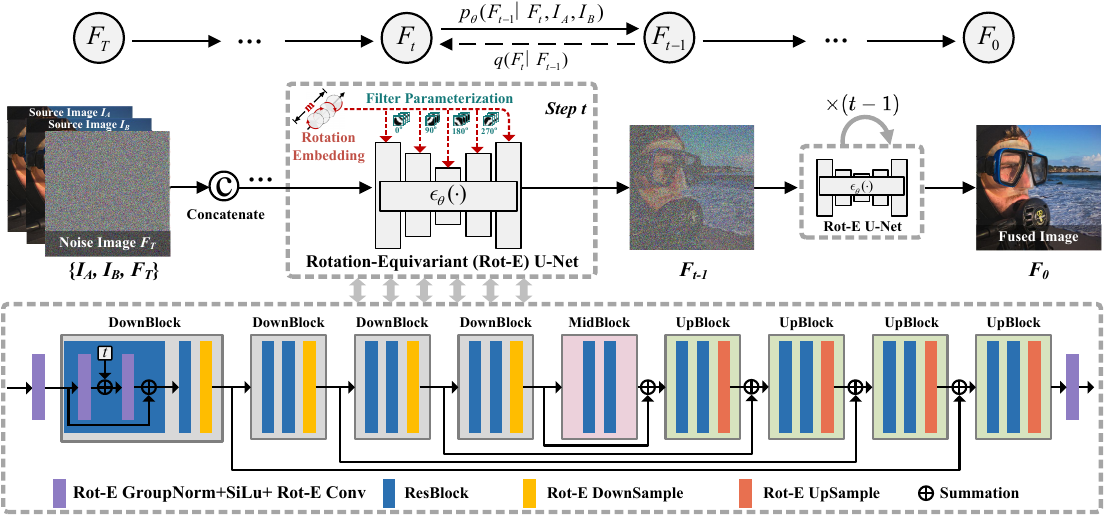}
  \caption{The MFIF framework based on the proposed \ours. 
          Rot-E Conv denotes the B-Conv~\cite{xie2025rotation}.
Due to efficient parameter sharing, each Rot-E convolution uses only $\tfrac{1}{m}$ of the parameters of a regular convolution, where $m$ denotes the equivariant group and is set to 4.
As a result, \ours is lightweight, reducing the number of parameters from 26.91M to 7.55M.
Moreover, under the $m=4$ rotation group, \ours theoretically achieves zero equivariance error (see Corollary~\ref{Corollary_1}).
}
  \label{fig:framewrk}
\end{figure*}

\textbf{Reverse Diffusion Process.}
The reverse process aims to recover $F_{t-1}$ from $F_t$ and is also modeled as a Markov chain:
\begin{equation}
p_\theta(F_{t-1}\mid F_t)
=
\mathcal{N}\!\left(F_{t-1};\mu_\theta(I_A,I_B,F_t,t),\, \sigma_t^2\mathbf{I}\right),
\label{eq:reverse}
\end{equation}
where the variance $\sigma_t^2$ is determined by the noise schedule, $I_A,I_B$ are source images, and the mean $\mu_\theta$ is parameterized via a conditional noise predictor $\epsilon_\theta(I_A,I_B,F_t,t)$:
\begin{equation}
    \begin{aligned}
\mu_\theta & =\frac{1}{\sqrt{\alpha_t}}\left(F_t-\frac{\beta_t}{\sqrt{1-\bar{\alpha}_t}} \epsilon_\theta\left(I_A, I_B, F_t, t\right)\right), \\
\sigma_t^2 & =\frac{1-\bar{\alpha}_{t-1}}{1-\bar{\alpha}_t} \beta_t.
\end{aligned}
\end{equation}

Given the deterministic nature of MFIF, we remove the stochastic noise term and perform sampling using only the mean of Eq.~\eqref{eq:reverse}, following~\cite{li2024fusiondiff}.

\begin{equation}
F_{t-1}
=
\frac{1}{\sqrt{\alpha_t}}
\left(
F_t
-
\frac{\beta_t}{\sqrt{1-\bar{\alpha}_t}}
\,\epsilon_\theta(I_A,I_B,F_t,t)
\right).
\label{eq:det_update}
\end{equation}

\textbf{Loss Function.}
We train the conditional noise predictor by minimizing the noise prediction error:
\begin{equation}
\mathcal{L}
=
\left\|
\epsilon-\epsilon_\theta(I_A,I_B,F_t,t)
\right\|_2,
\label{eq:loss}
\end{equation}
where $t \sim \mathrm{Uniform}(\{1,\dots,T\})$ and $F_t$ follows Eq.~\eqref{eq:reparam}.

\textbf{Rotation-Equivariant Denoising U-Net Network.}
As shown in \figurename~\ref{fig:framewrk}, the noise prediction U-Net network adopts a series of rotation-equivariant (Rot-E) modules to guarantee the entire equivariant property.
First, source images $I_A$ and $I_B$, along with their noisy image $F_T$, are concatenated, followed by a Rot-E layer (GN+SiLu+B-Conv~\cite{xie2025rotation}) for feature extraction.
The features are sequentially fed through four identical Rot-E DownBlocks to extract multi-resolution features.
Each DownBlock consists of two ResBlocks followed by a Rot-E downsampling layer.
Notably, time information is embedded into each ResBlock module.
Subsequently, the extracted features are refined via a MidBlock with two ResBlocks.
The decoder comprises four UpBlocks, each consisting of two ResBlocks and a Rot-E upsampling layer.
Before upsampling, features extracted at different resolutions by the encoder are aggregated to fully exploit multi-scale feature information.
Finally, the aggregated features are processed by a Rot-E layer (GN+SiLu+B-Conv) to generate the output.

\subsection{Theoretical Analysis of Equivariance}
\textbf{Definition of Rotation Equivariance.}
Following prior work~\cite{liu2025rotation}, a network (or mapping) is said to be rotation equivariant if rotating the input induces a corresponding rotation
of the output, without any additional change.
Mathematically, let $\Phi$ be a mapping from the input feature space to the output feature space, and let $S$ be a subgroup of rotation transformations,
\begin{equation}
S=\left\{
R_k=
\begin{bmatrix}
\cos \tfrac{2\pi k}{m} & -\sin \tfrac{2\pi k}{m} \\
\sin \tfrac{2\pi k}{m} & \cos \tfrac{2\pi k}{m}
\end{bmatrix}
\ \middle|\ k=0,1,\ldots,m-1
\right\}.
\end{equation}
Denote by $\pi_{\tilde{R}}^{I}$ the action of $\tilde{R}\in S$ on the input image
and by $\pi_{\tilde{R}}^{F}$ the corresponding action on output features.
Then $\Phi$ is equivariant with respect to $S$ if for any $\tilde{R}\in S$,
\begin{equation}
    \Phi\left[\pi_{\tilde{R}}^I\right](I)=\pi_{\tilde{R}}^F[\Phi](I),
\end{equation}
where $I$ denotes the input image, and $[\,\cdot\,]$ denotes function composition.
For details of the rotation operators
$\pi_{\tilde{R}}^{I}$ and $\pi_{\tilde{R}}^{F}$, please refer to Appendix~A.1.

Next, we analyze the equivariance error of the entire network.
\ours contains group convolutions, upsampling, downsampling, group normalization, and SiLU activation functions. 
Recent work~\cite{fu2024rotation} proves the equivariance of group convolution, and pointwise activation functions, such as SiLU, do not affect equivariance~\cite{weiler2019general}.
Therefore, we focus on the equivariance of the remaining modules and the overall \ours network.

\subsubsection{\textbf{Rotation Equivariance of Downsampling}}
Maxpooling is a commonly used downsampling operator. 
Other downsampling analyses are similar and are omitted here for brevity.
Formally, we can define the maxpooling operator $\Phi_\mathrm{MP}$ as follows:
\begin{equation}
[\Phi_\mathrm{MP}(F)](x,R)
=
\max_{(\hat{i},\hat{j})\in\Omega_x} F^{R}_{\hat{i} \hat{j}},
\end{equation}
where $x=[x_1,x_2]^{\mathsf{T}}\in\mathbb{R}^2$ is a continuous spatial location.
Let $i=\lfloor x_1/\delta\rfloor$ and $j=\lfloor x_2/\delta\rfloor$ be the grid
cell containing $x$ (mesh size $\delta$), and define
\begin{equation}
\Omega_x=\{(i,j),(i+1,j),(i,j+1),(i+1,j+1)\}.
\end{equation}

\begin{theorem}\label{th:downsampling}
Assume that a feature map $F$ is
discretized by the continuous function $e:\mathbb{R}^2\times S\to\mathbb{R}$, $|S|=m$, with mesh size $\delta$.
If for any $R\in S$ and $x\in\mathbb{R}^2$, the following condition is satisfied:
\begin{equation}
\|\nabla_x e(x,R)\|\le G,
\label{eq:gradient_bound_down}
\end{equation}
then for any $\tilde{R}\in S$, the following result holds:
\begin{equation}
\left|
\Phi_\mathrm{MP}\left(\pi^F_{\tilde{R}}(F)\right)(x,R)
-
\bigl[\pi^F_{\tilde{R}}(\Phi_\mathrm{MP}(F))\bigr](x,R)
\right|
\le
2\sqrt{2}\,G\delta.
\label{eq:main_result_down}
\end{equation}
\end{theorem}

Theorem~\ref{th:downsampling} shows that the equivariance error introduced by maxpooling goes to zero when the mesh size $\delta$ approaches zero.
See Appendix~A.2.

\subsubsection{\textbf{Rotation Equivariance of Upsampling}}
Bilinear interpolation is widely adopted for upsampling.
We can define the bilinear interpolation $\Phi_\mathrm{BI}$ as follows:
\begin{equation}
[\Phi_\mathrm{BI}(F)](x,R)
=
\sum_{v=1}^{2}\sum_{u=1}^{2}\lambda_{vu}(x)\,
F^{R}_{i+v-1,\,j+u-1},
\end{equation}
where $(i,j)$ is the grid cell containing $x$, and the weights satisfy
$\lambda_{vu}(x)\ge 0$ and $\sum_{v=1}^{2}\sum_{u=1}^{2}\lambda_{vu}(x)=1$.

\begin{theorem}\label{th:upsampling}
Assume the same setting as Theorem~\ref{th:downsampling}.
If for any $R\in S$ and
$x\in\mathbb{R}^2$, the following condition is satisfied:
\begin{equation}
\|\nabla_x e(x,R)\|\le G,
\label{eq:gradient_bound_up}
\end{equation}
then for any $\tilde{R}\in S$, the following result holds:
\begin{equation}
\left|
\Phi_\mathrm{BI}\left(\pi^F_{\tilde{R}}(F)\right)(x,R)
-
\bigl[\pi^F_{\tilde{R}}(\Phi_\mathrm{BI}(F))\bigr](x,R)
\right|
\le
2(\sqrt{2}+1)\,G\delta.
\label{eq:main_result_up}
\end{equation}
\end{theorem}

Theorem~\ref{th:upsampling} shows that the rotation equivariance error of bilinear interpolation mainly depends on the mesh size $\delta$. 
When the mesh size approaches zero, the equivariant error will also approach zero.
The proof is given in Appendix~A.3.

\subsubsection{\textbf{Rotation Equivariance of Normalization.}}
Group normalization, denoted by $\Phi_{\mathrm{GN}}$, is applied pointwise over $(x,R)$ and normalizes features across channel groups.
Let $\{\mathcal{C}_g\}_{g=1}^{N_g}$ partition $\{1,\ldots,n\}$.
For $F(x,R)\in\mathbb{R}^{n}$, let $\mu(x,R),\sigma^2(x,R)\in\mathbb{R}^{n}$ be the per-channel mean/variance obtained by computing statistics within each group and broadcasting them to channels in that group. Then
\begin{equation}
\Phi_{\mathrm{GN}}(F)(x,R)
=
\gamma \odot \frac{F(x,R)-\mu(x,R)}{\sqrt{\sigma^2(x,R)+\varepsilon}}
+\beta,
\label{eq:gn_def}
\end{equation}
where $\varepsilon>0$ is a small constant, and $\gamma,\beta\in\mathbb{R}^{n}$ are learnable parameters shared across $R\in S$.

\begin{theorem}\label{th:gn}
For any $\tilde{R} \in S$, group normalization is equivariant, i.e.,
for all $(x,R)\in\mathbb{R}^2\times S$, 
\begin{equation}
\Phi_\mathrm{GN}\left(\pi^F_{\tilde{R}}(F)\right)(x,R)
=
\bigl[\pi^F_{\tilde{R}}(\Phi_\mathrm{GN}(F))\bigr](x,R).
\label{eq:gn_equiv}
\end{equation}
\end{theorem}

Theorem~\ref{th:gn} shows that group normalization does not affect equivariance (see Appendix~A.4).

\subsubsection{\textbf{Rotation Equivariance of Entire ReDiffuse Network.}}
We next analyze the equivariance error of the entire network under discrete rotations.
\ours employs a U-Net backbone network, which can be defined as follows:
\begin{equation}
\mathrm{ReDiffuse}
=
\hat{\Upsilon}
\circ
\Phi_{\mathrm{BI}}^{(i)} \circ \cdots \circ \Phi_{\mathrm{BI}}^{(1)}
\circ
\hat{\mathcal{F}}
\circ
\Phi_{\mathrm{MP}}^{(i)} \circ \cdots \circ \Phi_{\mathrm{MP}}^{(1)}
\circ
\hat{\Psi},
\label{eq:unet_definition}
\end{equation}
where $\hat{\Psi}$, $\hat{\mathcal{F}}$, and $\hat{\Upsilon}$
denote the convolutional layers in the
input, middle, and output stages, respectively.
For conciseness, we omit SiLU and GN in Eq.~\eqref{eq:unet_definition}, as both do not affect the rotation equivariance of the network.
 
\begin{theorem}\label{th:enitre_network}
For an image $I \in \mathbb{R}^{h \times w \times n_0}$ and a $N$-layer $\mathrm{ReDiffuse}(\cdot)$, whose channel number of the $l$-th layer is $n_l$, rotation subgroup is $S \le O(2)$, $|S| = m$, and activation function is SiLU.
Denote the latent continuous function of the $c$-th channel of $I$ as
$r_c : \mathbb{R}^2 \to \mathbb{R}$,
and the latent continuous function of any convolutional filter
in the $l$-th layer is
$\phi_l : \mathbb{R}^2 \to \mathbb{R}$.
The following conditions are satisfied:
\begin{equation}
\begin{aligned}
& |r_c(x)| \le F_0,
\|\nabla_x r_c(x)\| \le G_0,
 \|\nabla_x^2 r_c(x)\| \le H_0, \\
& |\phi_l(x)| \le F_l,
\|\nabla_x \phi_l(x)\| \le G_l,
 \|\nabla_x^2 \phi_l(x)\| \le H_l, \\
& \phi_l(x)=0,
\forall \|x\| \ge (p+1)\delta/2 ,
\end{aligned}
\label{eq:unet_assumptions}
\end{equation}
where $p$ denotes the filter size, $\delta$ is the mesh size, $\nabla_x$ and $\nabla_x^2$ denote the operators of gradient and Hessian matrix, respectively.
Then, for discrete rotation angles $\theta_k = \frac{2k\pi}{m}$, $k=0,\ldots,m-1$, the following result is satisfied:
\begin{equation}
\left|
\mathrm{ReDiffuse}\bigl(\pi^I_{R_{\theta_k}}(I)\bigr)
-
\pi^I_{R_{\theta_k}}[\mathrm{ReDiffuse}](I)
\right|
\le
C_1 \delta,
\label{eq:unet_equivariance_error}
\end{equation}
where $C_1$ is a constant, as detailed in Appendix A.5.
\end{theorem}

For the most commonly used $m=4$ rotation group, it is easy to deduce that the proposed \ours theoretically achieves zero equivariance error.
We thus obtain the following corollary, whose proof is provided in Appendix~A.6.

\begin{corollary}\label{Corollary_1}
Under the same conditions as in Theorem~\ref{th:enitre_network} and $m=4$, the following result is satisfied:

\begin{equation}
\mathrm{ReDiffuse}\bigl(\pi^I_{R_{\theta_k}}(I)\bigr)=\pi^I_{R_{\theta_k}}[\mathrm{ReDiffuse}](I).
\label{eq:m=4}
\end{equation}

\end{corollary}

Moreover, we can further derive an upper bound on the equivariant error for arbitrary rotation angles, as given below.
\begin{corollary}\label{Corollary_2}
Under the same conditions as in Theorem~\ref{th:enitre_network},
for an arbitrary rotation angle $\theta \in [0,2\pi]$,
let $R_\theta$ denote the $\theta$-degree rotation matrix, then for any $\theta$ we have 
\begin{equation}
\left|
\mathrm{ReDiffuse}\bigl(\pi^I_{R_{\theta}}(I)\bigr)
-
\pi^I_{R_{\theta}}[\mathrm{ReDiffuse}](I)
\right|
\le
C_1 \delta +  C_2m^{-1}\delta ,
\end{equation}
where $C_1$ and $C_2$ are constants whose explicit expressions
are given in Appendix~A.7.
\end{corollary}

\begin{figure}[t]
  \begin{center}
    \centerline{\includegraphics[width=\linewidth, keepaspectratio]{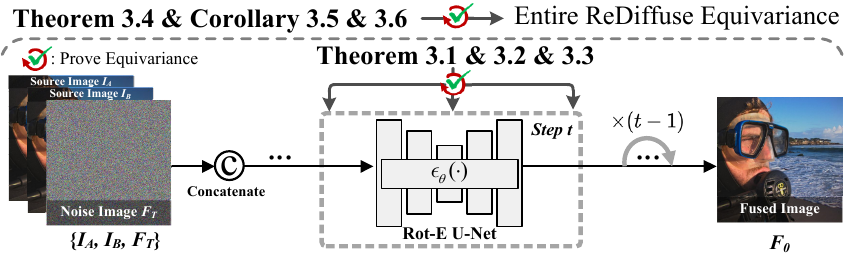}}
    \caption{
    Illustration of the proposed theoretical results and their correspondence with the network modules.
    }
    \label{fig:Theory}
  \end{center}
\end{figure}

Finally, to more intuitively and clearly illustrate the correspondence between the proposed theoretical results and the network modules, \figurename~\ref{fig:Theory} provides a schematic diagram.


\begin{table*}[!ht]
  \centering
   \caption{Main quantitative MFIF results. 
  The best and second-best values are highlighted in \colorbox{firstcolor}{red} and \colorbox{secondcolor}{blue}, respectively.
    }
   \resizebox{\linewidth}{!}{
    \begin{tabular}{lccccccccccccc}
\toprule
        \multirow{2}{*}{Methods}       & \multicolumn{6}{c}{\textbf{Lytro Multi-focus Fusion Dataset}}                                                                                                                                                                    &               & \multicolumn{6}{c}{\textbf{MFFW Multi-focus Fusion Dataset}}                                                                                                                                                                                                    \\ 
\cmidrule(l){2-7}\cmidrule(l){9-14}
           & Qabf$\uparrow$                            & QMI$\uparrow$                               & QG$\uparrow$                               & QP$\uparrow$                             & QE$\uparrow$                             & \small{MS-SSIM}$\uparrow$                              &               & Qabf$\uparrow$                            & QMI$\uparrow$                               & QG$\uparrow$                               & QP$\uparrow$                             & QE$\uparrow$                             & \small{MS-SSIM}$\uparrow$                               \\ 
\midrule
MFF-GAN  \hfill\graytext{[InFus2021]}         & 0.630                                     & 0.720                                    & 0.522                                     & 0.686                                    & 0.801 & 0.962                                  &           & 0.574                                     & 0.709                                   &0.441                                     & 0.489                                   &0.739                                    & 0.969                                  \\
SwinFusion  \hfill\graytext{[JAS2022]}         & 0.690                                     & 0.748                                     & 0.585                                     & 0.723                                    & 0.848 & 0.979                                    &           & 0.603                                     & 0.700                                     &0.463                                     & 0.505                                    &0.747                                     & 0.968                                  \\
ZMFF  \hfill\graytext{[InfFus2023]}         & 0.703                                    & 0.884                                     &0.631                                     & 0.785                                     & 0.891 & 0.992                                     &           & 0.666                                    & 0.766                                    & 0.564                                    & 0.647                                    &0.809                                    & 0.975                                     \\
FusionDiff  \hfill\graytext{[ESWA2024]}         & 0.711                                    & 0.903                                    &0.640                                    & 0.795                                     &0.856 & 0.991                                     &           & 0.659                                   & \cellcolor[rgb]{ .863,  .902,  .941}0.821                                  &0.574                                    & 0.654                                     &0.750                                    & 0.980                                    \\
CCF  \hfill\graytext{[NeurIPS2024]}        &0.491                                     & 0.799                                    & 0.411                                     & 0.587                                    & 0.476                                    & 0.949                                     &        &                           0.450         & 0.761                                     & 0.375                                    & 0.432                                    & 0.391                                   & 0.944                                      \\
Deep$\mathrm{\text{M}^{2}}$CDL  \hfill\graytext{[TPAMI2024]}   & 0.718                                     & 0.867                                    & 0.637                                   & 0.809                               & 0.896                                     & 0.991                                     &   & \cellcolor[rgb]{ .863,  .902,  .941}0.672 & 0.778                                     & 0.566                                     & \cellcolor[rgb]{ .863,  .902,  .941}0.666 & \cellcolor[rgb]{ .863,  .902,  .941}0.815                                     & \cellcolor[rgb]{ .863,  .902,  .941}0.982  \\
TC-MoA  \hfill\graytext{[CVPR2024]}      & \cellcolor[rgb]{ .863,  .902,  .941}0.740                                     & \cellcolor[rgb]{ .945,  .863,  .859}0.993                                     & \cellcolor[rgb]{ .863,  .902,  .941}0.689                                     &\cellcolor[rgb]{ .863,  .902,  .941}0.819                                    & \cellcolor[rgb]{ .863,  .902,  .941}0.902                                   & \cellcolor[rgb]{ .863,  .902,  .941}0.994                                    &     & 0.603                                   & 0.755                                     & 0.465 & 0.531                                    &0.772                                     &0.979                                     \\
ReFusion \hfill\graytext{[IJCV2025]}     & 0.698                                    & 0.888                                    & 0.613                                     & 0.767                                    & 0.871 & 0.990                                  &      & 0.655                                     & 0.814                                    & \cellcolor[rgb]{ .863,  .902,  .941}0.568                                     & 0.650                                     & 0.771                                    & 0.980                                      \\
VDMUFuse \hfill\graytext{[TIP2025]}    & 0.560 & 0.837 & 0.482 & 0.667                                     & 0.550                                    & 0.964                                     &    & 0.533                                     & 0.803                                     & 0.469                                     & 0.573                                     & 0.504                                     & 0.960                                      \\
Mask-DiFuser  \hfill\graytext{[TPAMI2025]}       & 0.600                                    & 0.736                                   &0.498                                    &0.677 & 0.662                                    & 0.955                                    &        & 0.559                                     & 0.713                                     & 0.437                                   & 0.537                                     & 0.589                                    & 0.949                                      \\
GIFNet \hfill\graytext{[CVPR2025]}       & 0.519                                   & 0.740                                    & 0.438                                     & 0.541                                     & 0.591                                    &0.943 &       & 0.451                                     & 0.696 & 0.366                                     & 0.385                                    & 0.493 & 0.932                                      \\
\textbf{ReDiffuse (Ours)} & \cellcolor[rgb]{ .945,  .863,  .859}0.750 &\cellcolor[rgb]{ .863,  .902,  .941}0.982 & \cellcolor[rgb]{ .945,  .863,  .859}0.701 & \cellcolor[rgb]{ .945,  .863,  .859}0.840 & \cellcolor[rgb]{ .945,  .863,  .859}0.906 & \cellcolor[rgb]{ .945,  .863,  .859}0.995 &  & \cellcolor[rgb]{ .945,  .863,  .859}0.708 &\cellcolor[rgb]{ .945,  .863,  .859}0.864 & \cellcolor[rgb]{ .945,  .863,  .859}0.644 &\cellcolor[rgb]{ .945,  .863,  .859}0.727 & \cellcolor[rgb]{ .945,  .863,  .859}0.832 & \cellcolor[rgb]{ .945,  .863,  .859}0.987  \\ 
\midrule
     \multirow{2}{*}{Methods}         & \multicolumn{6}{c}{\textbf{ MFI-WHU Multi-focus Fusion Dataset}}                                                                                                                                                                                                &               & \multicolumn{6}{c}{\textbf{Road-MF Multi-focus Fusion Dataset}}                                                                                                                                                                                                    \\ 
\cmidrule(l){2-7}\cmidrule(l){9-14}
               & Qabf$\uparrow$                            & QMI$\uparrow$                               & QG$\uparrow$                               & QP$\uparrow$                             & QE$\uparrow$                             & \small{MS-SSIM}$\uparrow$                              &               & Qabf$\uparrow$                            & QMI$\uparrow$                               & QG$\uparrow$                               & QP$\uparrow$                             & QE$\uparrow$                             & \small{MS-SSIM}$\uparrow$                               \\ 
\midrule
MFF-GAN  \hfill\graytext{[InFus2021]}         & 0.607                                     & 0.705                                   & 0.545                                     & 0.661                                    & 0.718 & 0.953                                  &           & 0.594                                     & 0.951                                   &0.483                                     & 0.693                                   &0.717                                   & 0.970                                  \\
SwinFusion  \hfill\graytext{[JAS2022]}         & 0.658                                     & 0.735                                     & 0.603                                    & 0.689                                   & 0.804 & 0.978                                    &           & 0.683                                     & 1.039                                    &0.580                                     & 0.719                                    &0.760                                     & 0.993                                  \\
ZMFF  \hfill\graytext{[InfFus2023]}         & 0.632                                    & 0.789                                    &0.594                                    & 0.663                                     &0.801 & 0.989                                     &           & 0.625                                    & 0.891                                  &0.543                                    & 0.718                                     &0.791                                    & 0.967                                    \\
FusionDiff  \hfill\graytext{[ESWA2024]}         & \cellcolor[rgb]{ .863,  .902,  .941}0.710                                    & \cellcolor[rgb]{ .863,  .902,  .941}0.982                                    &\cellcolor[rgb]{ .863,  .902,  .941}0.681                                    & \cellcolor[rgb]{ .863,  .902,  .941}0.752                                     &0.829 & 0.992                                    &           & \cellcolor[rgb]{ .863,  .902,  .941}0.705                                    & \cellcolor[rgb]{ .863,  .902,  .941}1.148                                  &\cellcolor[rgb]{ .863,  .902,  .941}0.661                                    & \cellcolor[rgb]{ .863,  .902,  .941}0.829                                     &\cellcolor[rgb]{ .863,  .902,  .941}0.826                                    & 0.993                                    \\
CCF  \hfill\graytext{[NeurIPS2024]}        &0.497                                     & 0.790                                   & 0.440                                    & 0.561                                    & 0.478                                    & 0.960                                    &        &                           0.510         & 0.977                                     & 0.447                                    & 0.537                                    & 0.406                                   &0.972                                      \\
Deep$\mathrm{\text{M}^{2}}$CDL  \hfill\graytext{[TPAMI2024]}   & 0.679                                     & 0.861                                    & 0.638                                   & 0.738                             & \cellcolor[rgb]{ .863,  .902,  .941}0.831                                     & \cellcolor[rgb]{ .863,  .902,  .941}0.993                                     &   & 0.662 & 1.044                                     &0.601                                    & 0.775 & 0.794                                     & 0.987  \\
TC-MoA  \hfill\graytext{[CVPR2024]}      & 0.656                                     & 0.926                                     & 0.648                                     &0.710                                   & 0.772                                   & 0.991                                    &     & 0.681                                   & 1.114                                    & 0.660 & 0.756                                    &0.802                                     &0.989                                     \\
ReFusion \hfill\graytext{[IJCV2025]}     & 0.683                                    & 0.875                                    &0.652                                    & 0.733                                    & 0.825 & \cellcolor[rgb]{ .863,  .902,  .941}0.993                                &      & 0.687                                    & 1.097                                    & 0.635                                     &0.804                          & 0.815                                    & \cellcolor[rgb]{ .863,  .902,  .941}0.994                                      \\
VDMUFuse \hfill\graytext{[TIP2025]}    & 0.561 & 0.820 & 0.510 & 0.651                                     & 0.578                                    & 0.978                                     &    & 0.618                                     & 1.058                                    & 0.566                                     &0.700                                     & 0.602                                     & 0.987                                     \\
Mask-DiFuser  \hfill\graytext{[TPAMI2025]}       & 0.557                                   & 0.718                                   &0.467                                    &0.623 & 0.648                                    & 0.937                                   &        & 0.579                                     & 0.820                                     &0.423                                   &0.610                                     &0.696                                   & 0.954                                      \\
GIFNet \hfill\graytext{[CVPR2025]}       & 0.439                                   & 0.665                                    & 0.406                                     & 0.464                                     & 0.471                                    &0.912 &       & 0.454                                     & 0.867 & 0.392                                     & 0.494                                    & 0.481 &0.939                                      \\
\textbf{ReDiffuse (Ours)} & \cellcolor[rgb]{ .945,  .863,  .859}0.725 &\cellcolor[rgb]{ .945,  .863,  .859}1.067 & \cellcolor[rgb]{ .945,  .863,  .859}0.718 & \cellcolor[rgb]{ .945,  .863,  .859}0.759 & \cellcolor[rgb]{ .945,  .863,  .859}0.844 & \cellcolor[rgb]{ .945,  .863,  .859}0.996 &  & \cellcolor[rgb]{ .945,  .863,  .859}0.717 &\cellcolor[rgb]{ .945,  .863,  .859}1.180 & \cellcolor[rgb]{ .945,  .863,  .859}0.695 &\cellcolor[rgb]{ .945,  .863,  .859}0.857 & \cellcolor[rgb]{ .945,  .863,  .859}0.832 & \cellcolor[rgb]{ .945,  .863,  .859}0.996  \\ 
\bottomrule
    \end{tabular}
    }
  \label{tab:quantitative}
\end{table*}%

\section{Experiments}

\subsection{Experimental Setup}
\textbf{Datasets. }
We train and validate the proposed \ours on the real-world MFIF dataset Real-MFF~\cite{zhang2020real}, which contains 600 image pairs for training and 110 image pairs for validation.
For evaluation, we conduct experiments on four widely used MFIF benchmarks, including Lytro~\cite{nejati2015multi}, MFFW~\cite{xu2020towards}, MFI-WHU~\cite{zhang2021mff}, and Road-MF~\cite{li2024samf}.
Among them, Lytro is a widely used MFIF dataset with 20 image pairs. 
MFFW contains 13 image pairs with more complex scenes and stronger defocus spread effects. 
MFI-WHU contains 120 image pairs. 
Road-MF is a road-scene MFIF dataset with 80 image pairs.

\noindent\textbf{Metrics.} We adopt six commonly used metrics to quantitatively evaluate fusion results from multiple perspectives, including assessment of blended features (Qabf), normalized mutual information (QMI), gradient-based index (QG), phase consistency (QP), edge-based index (QE), and multi-scale structural similarity (MS-SSIM).
It is worth noting that, due to the absence of ground-truth images in the MFIF dataset, we follow~\cite{zhao2023cddfuse, Zhao_2023_ICCV} to compute MS-SSIM. Specifically, we first compute the MS-SSIM between each source image and the fused image, and then take the average of the resulting values.
All metrics follow the principle that higher values indicate better performance.
Detailed definitions are provided in~\cite{liu2024rethinking}.

\noindent\textbf{Comparison Methods.} We compare our \ours with representative MFIF approaches spanning GAN, CNN, Transformer, and diffusion-based methods, including MFF-GAN~\cite{zhang2021mff}, SwinFusion~\cite{ma2022swinfusion}, ZMFF~\cite{hu2023zmff}, FusionDiff~\cite{li2024fusiondiff}, CCF~\cite{cao2024conditional}, Deep$\mathrm{\text{M}^{2}}$CDL~\cite{10323520}, TC-MoA~\cite{zhu2024task}, ReFusion~\cite{bai2025refusion}, VDMUFuse~\cite{shi2024vdmufusion}, Mask-DiFuser~\cite{tang2025mask}, and GIFNet~\cite{cheng2025one}.
To ensure fairness, all competing methods used the official provided code and weights.

\noindent\textbf{Implementation Details.}
All experiments are conducted on an NVIDIA V100 GPU.
Our proposed \ours is trained for 10,000 epochs with a batch size of 32 using the Adam optimizer.
The initial learning rate is set to 0.0002 and decayed by a factor of 0.99 every 1,000 epochs.
For the Rot-E U-Net, the initialized channel is set to 64, and the number of groups of GroupNorm is 16. 
For more details, please see Appendix B.

\begin{figure*}[t]
\centering
  \includegraphics[width=\linewidth]{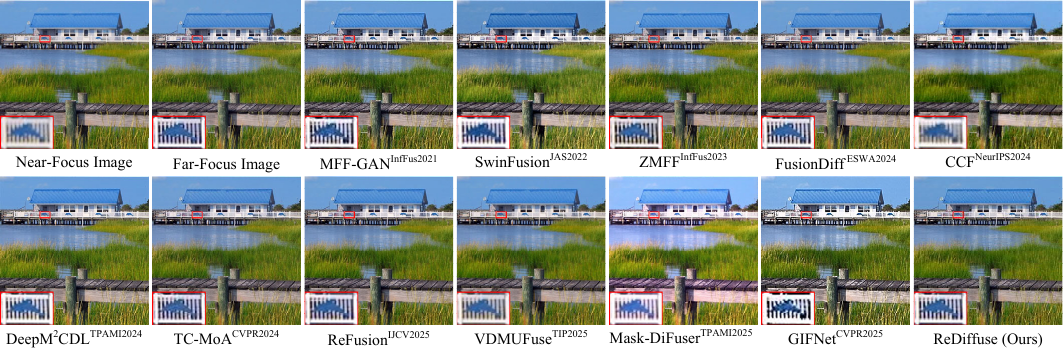}
  \caption{Visual comparison for ``MFFW-8'' in MFFW dataset, with the demarcated areas zoomed in 5 times for easy observation.}
  \label{fig:fusion_results}
\end{figure*}

\begin{figure*}
\centering
  \includegraphics[width=\linewidth]{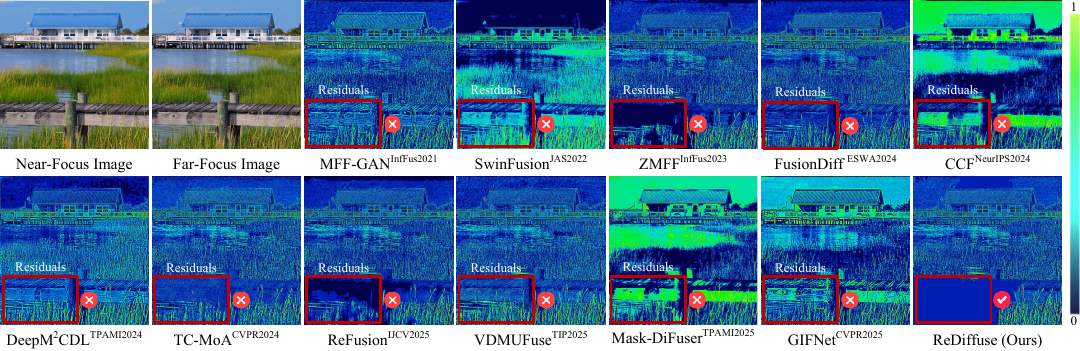}
  \caption{Visualization of difference images.
These images are obtained by subtracting the near-focus source image from the fused image.
\textbf{Less residual information indicates better fusion quality}.
  }
  \label{fig:diff}
\end{figure*}

\begin{figure*}
\centering
  \includegraphics[width=0.98\linewidth]{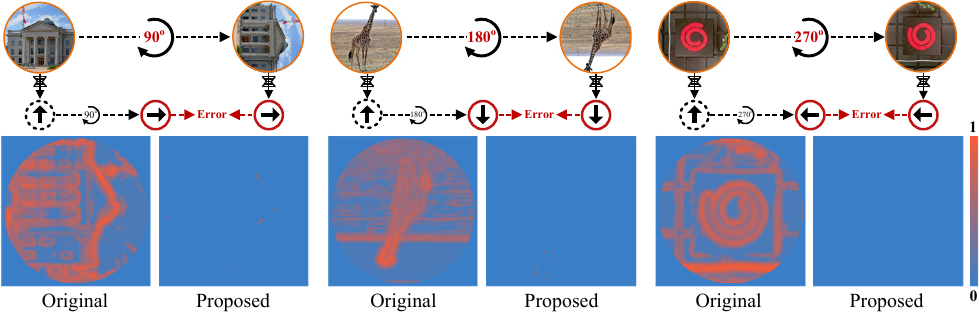}
  \caption{Illustration of rotation equivariant errors, defined as
$\left|
\Phi\!\left[\pi_{\tilde{R}}^I\right](I)
-
\pi_{\tilde{R}}^F\!\left[\Phi\right](I)
\right|$,
where $\Phi$ denotes the original model (without Rot-E) or the proposed \ours model.
Three images from the MFI-WHU dataset are evaluated under rotation angles of
$90^\circ$, $180^\circ$, and $270^\circ$.
\textbf{Please zoom for a better view.}
}
  \label{fig:error}
\end{figure*}

\subsection{Main Results}

\textbf{Quantitative Comparison.} \tablename~\ref{tab:quantitative} reports quantitative results on four widely adopted MFIF benchmarks across 11 representative methods.
The results indicate that \ours consistently achieves the best performance across nearly all evaluation metrics.
In particular, \ours outperforms the second-best method by margins of 2.63\%, 3.89\%, 6.64\%, 4.01\%, 1.21\%, and 0.28\% on Qabf, QMI, QG, QP, QE, and MS-SSIM, respectively, averaged over the four datasets.

It should be noted that rotation equivariance has a negligible impact on inference speed. 
Due to space limitations, we provide the complete runtime analysis in Appendix~C.1.

\noindent\textbf{Qualitative Comparison.} \figurename~\ref{fig:fusion_results} presents the visual results of all fusion methods for a pair of source images from the MFFW dataset. 
With the help of rotation equivariance, the proposed \ours achieves significant improvements in visual quality.
The enlarged red regions provide intuitive visual evidence.
As can be observed, \ours yields fusion results with superior overall clarity, such as more natural dolphin contours.
Moreover, \ours effectively preserves the original orientation consistency of the input images, enabling more faithful restoration of vertical line structures.
Due to space limitations, \textit{\textbf{more visualizations and downstream object detection}} results are provided in Appendix~C.2 and~C.3.

\noindent\textbf{Difference Image Comparison.}
To facilitate more intuitive comparison, we visualize difference images 
obtained by subtracting the source images from the fused results.
Intuitively, these residual maps reveal the extent to which useful information from the source images is lost during fusion: smaller residual responses generally correspond to better preservation of source content. As illustrated in \figurename~\ref{fig:diff}, \ours exhibits weaker residuals and fewer prominent response regions, indicating that it is more effective at retaining important structures, textures, and intensity information from the source images. This further confirms the advantage of \ours in preserving source information during the fusion process.

\subsection{Visualization Equivariance Errors}
\figurename~\ref{fig:error} illustrates the rotation equivariance error maps of the original model and the proposed \ours under rotation angles of 90$^\circ$, 180$^\circ$, and 270$^\circ$, evaluated after 1,000 training epochs. 
The original model exhibits pronounced rotation equivariance errors under rotational transformations. 
In contrast, \ours consistently produces substantially smaller equivariance errors (close to zero), demonstrating a marked improvement in rotation equivariance.
Notably, \textit{these results are highly consistent with the theoretical analysis presented in Theorem~\ref{th:enitre_network}}, providing further experimental validation of the soundness and effectiveness of the proposed theory.

\subsection{Generalization Verification}
We further evaluate the generalization ability of the proposed method on three representative diffusion-based fusion models, namely FusionDiff, VDMUFuse, and Mask-DiFuser, by replacing their original architectures with our rotation-equivariant design. 
As shown in \tablename~\ref{tab:generalization}, consistent performance gains are achieved.

\begin{table}
\centering
\caption{Generalization performance of the proposed rotation-equivariant method, evaluated on MFFW dataset.}
\label{tab:generalization}
\resizebox{\linewidth}{!}{
\begin{tabular}{lcccccc} 
\toprule
Method               & Mode & Qabf$\uparrow$ & QMI$\uparrow$    & QG$\uparrow$    & QP$\uparrow$  & QE$\uparrow$  \\ 
\midrule
\multirow{2}{*}{\small{FusionDiff}}       & Original  & 0.66 &0.82 & 0.57 & 0.65 & 0.75  \\
             & Proposed & \cellcolor[rgb]{ .945,  .863,  .859}0.69 & \cellcolor[rgb]{ .945,  .863,  .859}0.84 & \cellcolor[rgb]{ .945,  .863,  .859}0.60 & \cellcolor[rgb]{ .945,  .863,  .859}0.69 & \cellcolor[rgb]{ .945,  .863,  .859}0.78  \\
\midrule

\multirow{2}{*}{\small{VDMUFuse}}       & Original& 0.53 & 0.80 & 0.47 & 0.57 & 0.50  \\
              & Proposed & \cellcolor[rgb]{ .945,  .863,  .859}0.57 &\cellcolor[rgb]{ .945,  .863,  .859}0.82 & \cellcolor[rgb]{ .945,  .863,  .859}0.53 & \cellcolor[rgb]{ .945,  .863,  .859}0.61 & \cellcolor[rgb]{ .945,  .863,  .859}0.54  \\
\midrule

\multirow{2}{*}{\small{Mask-DiFuser}}       & Original& 0.56 &0.71 & 0.44 & 0.54 & 0.59  \\
              & Proposed  & \cellcolor[rgb]{ .945,  .863,  .859}0.59 &\cellcolor[rgb]{ .945,  .863,  .859}0.73 & \cellcolor[rgb]{ .945,  .863,  .859}0.48 & \cellcolor[rgb]{ .945,  .863,  .859}0.56 & \cellcolor[rgb]{ .945,  .863,  .859}0.61  \\
\bottomrule
\end{tabular}
}
\end{table}

\begin{table}
\centering
\caption{Ablation study on the effectiveness and superiority of rotation equivariance, evaluated on the MFFW dataset.}
\label{tab:ablation}
\setlength{\tabcolsep}{1.4mm}
\resizebox{\linewidth}{!}{
\begin{tabular}{ccccccc} 
\toprule
               &  Configurations & Qabf$\uparrow$ & QMI$\uparrow$    & QG$\uparrow$    & QP$\uparrow$  & QE$\uparrow$  \\ 
\midrule
I       & w/o rotation equivariance  & 0.67 &0.81 & 0.59 & 0.67 & 0.76  \\
II       & G-Conv $\to$ B-Conv   & 0.68 &0.81 & 0.60 & 0.69 & 0.81  \\
III       & F-Conv $\to$ B-Conv& 0.70 &0.84 & 0.61 & 0.71 & 0.80  \\
\midrule  
  & Ours& \cellcolor[rgb]{ .945,  .863,  .859}0.71 & \cellcolor[rgb]{ .945,  .863,  .859}0.86 & \cellcolor[rgb]{ .945,  .863,  .859}0.64 & \cellcolor[rgb]{ .945,  .863,  .859}0.73 & \cellcolor[rgb]{ .945,  .863,  .859}0.83  \\
\bottomrule
\end{tabular}
}
\end{table}

\subsection{Ablation Studies}
We conducted two ablation studies to validate the effectiveness and rationality of rotation equivariance.

\noindent\textbf{Effectiveness of Rotation-Equivariant.}
In Exp. I, we remove all equivariant modules. 
A consistent performance degradation is clearly observed, indirectly demonstrating the effectiveness of rotation equivariance, as shown in \tablename~\ref{tab:ablation}.

\noindent\textbf{Different Filter Parameterization Methods.}
In Exps. II and III, we compare different filter-parameterized equivariant convolution methods.
We replace the B-Conv~\cite{xie2025rotation} used in this work with G-Conv~\cite{weiler2019general} and F-Conv~\cite{xie2023Fourier}, respectively.
Results in \tablename~\ref{tab:ablation} show that our B-Conv achieves performance gains across all evaluated benchmarks, indicating its superiority.

\begin{figure}
  \begin{center}
    \centerline{\includegraphics[width=\linewidth, keepaspectratio]{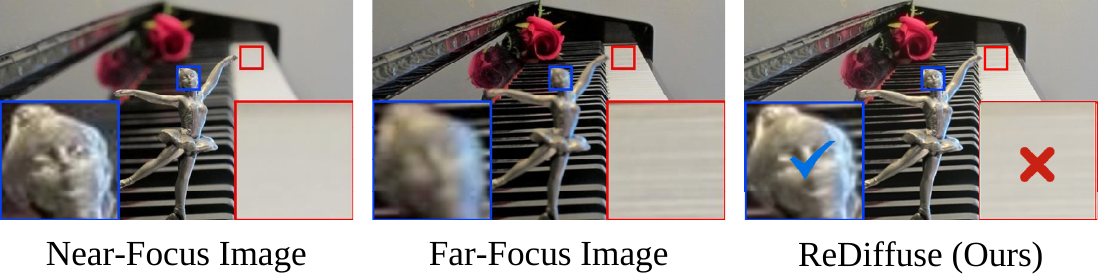}}
    \caption{
    Visualization of a failure case of \ours on the challenging MFFW dataset.
    }
    \label{fig:case}
  \end{center}
\end{figure}

\subsection{Failure Case Analysis}
In this section, we present a representative failure case to better understand the limitations of our \ours and gain insights into potential directions for further improvement.

As shown in \figurename~\ref{fig:case}, when the input images are affected by a strong defocus spread effect, the proposed \ours may make incorrect structural selections in some regions, failing to preserve the correct texture details from the in-focus focal plane in the fused image. 
This is a typical case of failure to maintain equivariance.

This issue can be attributed to the severe destruction of local texture structures caused by strong defocus spread, which hinders the model from correctly selecting and fusing structural information based solely on local features. 
According to Corollary~\ref{Corollary_2}, the equivariance error is jointly determined by the local derivatives $G_0$ and $H_0$ and the grid size $\delta$. 
Under stronger defocus spread, local structures become unstable and the derivative estimates deviate further from the true geometric patterns.
This increases the equivariance error and may lead to incorrect fusion results.

To alleviate this issue, exploiting non-local correlations~\cite{wang2018non} may be a worthwhile direction for future research. 
Non-local correlations help capture complementary information across different regions, better constraining structural selection in severely defocused areas and promoting more structurally consistent fusion results.

\section{Conclusion}
In this paper, we emphasize the key insight that under defocused conditions, existing diffusion models for MFIF are prone to violating the rotational consistency of geometric structures, introducing unintended artifacts in the generated images.
To address this, we propose \ours, which achieves rotation equivariance by carefully reconstructing the diffusion architecture.
This design ensures that fused results faithfully preserve the original orientation and structural consistency of geometric patterns in out-of-focus images.
Theoretical analysis proves that the rotation equivariance error of the proposed \ours can be made arbitrarily close to zero. 
Extensive experiments on four MFIF datasets further demonstrate the superiority of \ours.

\begin{acks}
This work was supported by Fundamental and Interdisciplinary Disciplines Breakthrough Plan of the Ministry of Education of China (JYB2025XDXM116), National Natural Science Foundation of China (No. 62137002, 62293553, 62450005, 62477036), the Fundamental Research Funds for the Central Universities (xzy022025037), and Xi'an  Jiaotong University City College Research Project (No. 2024Y01). 
\end{acks}

\bibliographystyle{ACM-Reference-Format}
\bibliography{main}







\end{document}